# GPT-4 passes most of the 297 written Polish Board Certification Examinations

Jakub Pokrywka[1], Jeremi I. Kaczmarek[1,2], Edward J. Gorzelańczyk[3,4,5]

[1]Adam Mickiewicz University, Poznań, Poland

[2]Poznań University of Medical Sciences, Poznań, Poland

[3]Department of Theoretical Basis of BioMedical Sciences and Medical Informatics, Nicolaus Copernicus University, Collegium Medicum, Bydgoszcz, Poland

[4]Institute of Philosophy, Kazimierz Wielki University, Bydgoszcz, Poland

[5]The Society for the Substitution Treatment of Addiction "Medically Assisted Recovery", Bydgoszcz, Poland

**Correspondence**:
Jakub Pokrywka, PhD
Wydział Matematyki i Informatyki
Uniwersytet im. Adama Mickiewicza
ul. Uniwersytetu Poznańskiego 4
61-614 Poznań, Poland

jakub.pokrywka@amu.edu.pl

jakubpokrywka@gmail.com

// Abstract

*Introduction:* Recently, the effectiveness of Large Language Models (LLMs) has increased rapidly, allowing them to be used in a great number of applications. However, the risks posed by the generation of false information through LLMs significantly limit their applications in sensitive areas such as healthcare, highlighting the necessity for rigorous validations to determine their utility and reliability. To date, no study has extensively compared the performance of LLMs on Polish medical examinations across a broad spectrum of specialties on a very large dataset.

*Objectives:* This study evaluated the performance of three Generative Pretrained Transformer (GPT) models on the Polish Board Certification Exam (Państwowy Egzamin Specjalizacyjny, PES) dataset, which consists of 297 tests.

*Methods:* We developed a software program to download and process PES exams and tested the performance of GPT models using OpenAI Application Programming Interface.

*Results:* Our findings reveal that GPT-3.5 did not pass any of the analyzed exams. In contrast, the GPT-4 models demonstrated the capability to pass the majority of the exams evaluated, with the most recent model, gpt-4-0125, successfully passing 222 (75%) of them. The performance of the GPT models varied significantly, displaying excellence in exams related to certain specialties while completely failing others.

*Conclusions:* The significant progress and impressive performance of LLM models hold great promise for the increased application of AI in the field of medicine in Poland. For instance, this advancement could lead to the development of AI-based medical assistants for healthcare professionals, enhancing the efficiency and accuracy of medical services.

# Introduction

## LLMs' development

In recent years, the rapidly evolving field of Artificial Intelligence (AI), particularly Generative Artificial Intelligence (GAI), has attracted considerable attention from the academic community and the public alike. Neural Language Models (LMs) proposed in the year 2000[1], soon proved their effectiveness in comprehending words semantic in architectures like Word2Vec[2], GloVe [3], fastText[4]. Shortly after the introduction of neural attention mechanism [5], it was used in transformer architecture[6]. Among the most prominent examples of transformers is the Generative Pretrained Transformer (GPT) family[7], [8], [9], [10] developed by OpenAI. Large Language Models (LLMs), such as GPT, can generate pages of coherent text within seconds, staying relevant to the context and exhibiting an understanding of complex concepts. Their foundation rests on machine learning, a method for training AI models by immersing them in extensive datasets covering various topics.

Considering that the datasets employed in the development of GPT encompassed professional medical literature and additional related content, given the premise that "understanding" is a trait conceivable within AI capabilities, a reasonable hypothesis would be that these models possess an extent of medical knowledge and display a degree of clinical reasoning[11], [12]. Nonetheless, it is crucial to recognize the possibility that the training data, including its span across medical sciences, may not have been entirely accurate. This is one of the contributing factors as to why GPT and other LLMs can produce "hallucinations" – outputs that, while often coherent and grammatically sound, are factually incorrect[13], [14], [15]. This limitation restricts their use in certain applications, notably in the medical sector where the accuracy of information is critical[16].

## LLMs' performance on medical tests

Since there are no foolproof methods to prevent hallucinations, it is crucial to validate LLMs' knowledge and reasoning abilities before they are approved for use. Their performance in various tasks is a growing interest[17], [18]. In a medical context, researchers have challenged GPT, among other models, with various assessments. These include licensure or board certification examinations, such as the United States Medical Licensing Examination (USMLE)[19], the European Exam in Core Cardiology[20], tests designed for Family Medicine residents[21], Medical Specialty Exams[22], orthopedics[23], ophthalmology[24], neurosurgery[25], [26].

While LLMs' performance in medical examinations has already been extensively studied, an important remark is that most of the analyses carried out so far regarded tests for English-speaking examinees. The language used significantly influences these models' output, making it essential to evaluate their abilities in different linguistic contexts as well. Beyond English, LLMs were assessed in medical tests in languages such as Chinese [27], [28]  Korean[29], Japanese[30], Spanish[31] and others.

## LLMs' performance on Polish medical tests

Several investigations employing various models and methodological approaches have been conducted to evaluate GPT's performance on medical examinations used in Poland. An experiment regarding the Medical Final Examination (Lekarski Egzamin Końcowy, or LEK, in Polish), necessary for obtaining a medical license in Poland, was recently reported[32]. Additionally, analyses targeting specific medical specialties, were made regarding the board certification exams in several fields, including cariology[33], cardiology[34], radiology & diagnostic imaging[35] and internal medicine[36]. In the latter, the GPT-3.5 model did not achieve the minimum score required for passing. However, it is important to consider that since this study was carried out, more sophisticated and up-to-date models of GPT have been released. Investigations into the accuracy of LLMs in responding to queries requiring medical knowledge and reasoning, which utilized the latest versions of GPT, support the proposition that these recently introduced models significantly outperform their predecessors[32].

Regardless of the language used, a predominant approach of the studies to date was to evaluate LLMs' performance on tests that assessed general medical knowledge, such as medical licensing examinations, or on tests that focused on a specific medical domain, e.g., board certification exams. To our best knowledge, there are no reports of experiments that applied consistent methodology in comparing LLMs' performance across a broad array of specialist-level medical exams. Especially, a study assessing different GPT models' performance across various PES exams on a large scale has not been conducted so far.

The Państwowy Egzamin Specjalizacyjny (PES), which translates to English as Polish Board Certification Exam in English, serves as a crucial assessment for medical practitioners in Poland, marking the culmination of their specialization process. It is composed of two parts: a written test with multiple-choice questions and an oral examination. Its primary objective is to evaluate the proficiency and expertise acquired during specialized training. The successful completion of the PES, in conjunction with the requisite courses and qualifying training periods, is obligatory for medical doctors in Poland to gain official recognition as specialists in their respective fields, thereby granting them the autonomy to practice their specialty independently.

We believe that truthful LLMs may be a great tool for serving medical education [37], [38], [39], assisting healthcare providers [40], [41], creating documentation [42], and other medical-related tasks. The motivation of this study is to evaluate, whether the GPT models may possess the essential knowledge to be effective in these areas considering the Polish language.

# Objectives

Our aim is to assess the performance of three GPT models (gpt-3.5-turbo, gpt-4-0613, and gpt-4-0125-preview) on the written component of the PES, covering 57 medical specialties and utilizing a database of 297 exams. By doing so, we wanted to expand the available data on GPT's performance in the PES onto specialties that had not been previously studied. Furthermore, we wished to compare the results across various medical domains to verify whether the models perform equally in all specialties. Finally, we wanted to follow up on the studies that were already conducted by applying the most up-to-date GPT models. This study provides a comprehensive comparison of AI performance in specialist-level testing across an extensive spectrum of medical disciplines. It is the first of its magnitude regarding medical examinations in the Polish language.

# Methods

PES exams were published by the Polish Supreme Medical Chamber (Naczelna Izba Lekarska, or NIL in Polish). The exams conducted from 2018 to 2022 in 57 medical and dental specialties were made publicly available on their internet sites (https://nil.org.pl/aktualnosci/8012-sukces-samorzadu-lekarskiego-nil-udostepnia-pytania-z-pes, https://nil.org.pl/aktualnosci/8043-kolejne-pytania-z-pes-udostepnione%C2%A0). The exams were published as zip catalogs, each containing several PDF files of tests and separate PDFs of corresponding correct answers.

We created a web scraper to download all the tests from the provided links. The web scraper was developed using Python 3 programming language and utilized the BeautifulSoup Python library. This approach enabled us to acquire all the published files automatically.

Unfortunately, PDF files of all the tests from 2021-2022 and several from 2018-2020 did not include text layers, making it impossible to retrieve the data automatically. That is why we did not include these exams in our analysis.

The test and answer files with a text layer were extracted using the PyPDF2 library in Python 3 and converted to JSON text files. Then, we utilized the OpenAI Application Programming Interface (API) (https://platform.openai.com/docs/overview) to script queries to GPT models for answers. It allowed us to automate querying the GPT models, making one call per question. Consequently, the GPT models were exposed to only one question at a time, followed by a model reset. We set the GPT inference temperature parameter to 0, indicating that the GPT models should operate with factual consistency and determinism (https://platform.openai.com/docs/guides/text-generation/how-should-i-set-the-temperature-parameter). We tested three models (https://platform.openai.com/docs/models/overview):

- gpt-3.5-turbo (training data up to September 2021)
- gpt-4-0613 (training data up to September 2021)
- gpt-4-0125-preview (training data up to December 2023)

We used the following prompt in Polish, which is the language of all tests:

*Twoje zadanie to udzielenie odpowiedzi na test medyczny dla lekarzy. Spośród wszystkich odpowiedzi wybierz tylko jedną. Odpowiedz tylko i wyłącznie jedną literą.*

We translated the prompt for the non-Polish speaking readers:

*Your task is to provide an answer to a medical test question for doctors. Choose only one answer from all the options. Respond with only one letter.*

The accuracy assessment and further analysis were conducted using Pandas and Scikit-Learn libraries.

**Specialties A-D**

| Exam | Result [%] | | |
|---|---|---|---|
| | gpt-3.5-turbo | gpt-4-0613 | gpt-4-0125-preview |
| **Allergology** *Alergologia* | | | |
| 2018 S | 43,33 | 66,67 | 64,17 |
| 2018 A | 38,33 | 65,00 | 71,67 |
| 2019 S | 42,50 | 68,33 | 67,50 |
| 2019 A | 28,33 | 59,17 | 57,50 |
| 2020 S | 36,67 | 62,50 | 64,17 |
| 2020 A | 33,33 | 70,00 | 63,33 |
| **Anesthesiology & critical care** *Anestezjologia i intensywna terapia* | | | |
| 2018 S | 41,67 | 57,50 | 60,83 |
| 2018 A | 36,67 | 72,50 | 75,00 |
| 2019 S | 43,33 | 68,33 | 70,83 |
| 2020 A | 40,00 | 70,83 | 72,50 |
| **Angiology** *Angiologia* | | | |
| 2019 A | 25,83 | 60,00 | 60,83 |
| 2020 S | 33,33 | 55,83 | 57,50 |
| 2020 A | 35,00 | 59,17 | 60,00 |
| **Balneology & physical medicine** *Balneologia i medycyna fizykalna* | | | |
| 2018 S | 31,67 | 54,17 | 49,17 |
| 2018 A | 25,00 | 38,33 | 48,33 |
| 2019 S | 34,17 | 49,17 | 53,33 |
| 2019 A | 35,00 | 47,50 | 46,67 |
| 2020 S | 37,50 | 45,83 | 49,17 |
| 2020 A | 28,33 | 48,33 | 54,17 |
| **Cardiac surgery** *Kardiochirurgia* | | | |
| 2019 S | 35,00 | 65,83 | 70,83 |
| 2020 S | 42,50 | 70,83 | 70,83 |
| **Cardiology** *Kardiologia* | | | |
| 2018 S | 30,00 | 59,17 | 56,67 |
| 2018 A | 35,00 | 55,00 | 60,00 |
| 2019 S | 35,83 | 54,17 | 61,67 |
| 2019 A | 37,50 | 58,33 | 66,67 |
| 2020 S | 40,00 | 50,00 | 60,83 |
| 2020 A | 28,33 | 65,00 | 61,67 |
| **Clinical oncology** *Onkologia kliniczna* | | | |
| 2018 S | 33,33 | 56,67 | 63,33 |
| 2018 A | 38,33 | 61,67 | 65,00 |
| 2019 S | 31,67 | 60,00 | 62,50 |
| 2019 A | 25,83 | 52,50 | 56,67 |
| 2020 S | 35,83 | 66,67 | 62,50 |
| 2020 A | 36,67 | 60,00 | 66,67 |
| **Clinical transplantology** *Transplantologia kliniczna* | | | |
| 2018 S | 34,17 | 61,67 | 65,83 |
| 2018 A | 32,50 | 56,67 | 63,33 |
| 2019 S | 44,17 | 60,83 | 60,00 |
| 2019 A | 34,17 | 58,33 | 61,67 |
| 2020 S | 40,00 | 64,17 | 65,83 |
| 2020 A | 33,33 | 67,50 | 70,83 |
| **Conservative dentistry** *Stomatologia zachowawcza* | | | |
| 2018 S | 29,17 | 46,67 | 44,17 |
| 2018 A | 25,83 | 46,67 | 50,83 |
| 2019 A | 28,33 | 49,17 | 41,67 |
| 2020 S | 23,33 | 50,83 | 51,67 |
| 2020 A | 28,33 | 49,17 | 50,83 |
| **Dental surgery** *Chirurgia stomatologiczn* | | | |
| 2018 S | 31,67 | 46,67 | 60,00 |
| 2018 A | 27,50 | 55,83 | 55,00 |
| 2019 S | 27,50 | 40,83 | 51,67 |
| 2019 A | 40,83 | 44,17 | 55,83 |
| 2020 S | 23,33 | 41,67 | 43,33 |
| 2020 A | 23,33 | 50,00 | 48,33 |

**Specialties D-G**

| Exam | Result [%] | | |
|---|---|---|---|
| | gpt-3.5-turbo | gpt-4-0613 | gpt-4-0125-preview |
| **Dentistry (pediatric)** *Stomatologia dziecięca* | | | |
| 2018 S | 21,67 | 54,17 | 50,00 |
| 2018 A | 32,50 | 41,67 | 53,33 |
| 2019 S | 33,33 | 52,50 | 55,00 |
| 2019 A | 30,83 | 45,83 | 55,00 |
| 2020 S | 27,50 | 47,50 | 46,67 |
| **Dermatology & venereology** *Dermatologia i wenerologia* | | | |
| 2018 S | 44,17 | 74,17 | 79,17 |
| 2018 A | 44,17 | 67,50 | 73,33 |
| 2019 S | 45,83 | 68,33 | 77,50 |
| 2019 A | 35,00 | 73,33 | 75,83 |
| 2020 S | 45,00 | 75,00 | 78,33 |
| 2020 A | 47,50 | 57,50 | 67,50 |
| **Diabetology** *Diabetologia* | | | |
| 2018 S | 38,33 | 66,67 | 62,50 |
| 2018 A | 35,00 | 60,00 | 60,00 |
| 2019 S | 36,67 | 50,83 | 52,50 |
| 2019 A | 37,50 | 61,67 | 60,00 |
| 2020 S | 30,00 | 53,33 | 57,50 |
| 2020 A | 38,33 | 60,83 | 57,50 |
| **Emergency medicine** *Medycyna ratunkowa* | | | |
| 2018 S | 45,00 | 64,17 | 66,67 |
| 2018 A | 49,17 | 63,33 | 69,17 |
| 2019 S | 44,17 | 63,33 | 69,17 |
| 2019 A | 44,17 | 65,83 | 75,00 |
| 2020 S | 36,67 | 63,33 | 64,17 |
| 2020 A | 50,83 | 75,00 | 76,67 |
| **Endocrinology** *Endokrynologia* | | | |
| 2018 S | 30,00 | 59,17 | 57,50 |
| 2018 A | 34,17 | 52,50 | 51,67 |
| 2019 S | 25,00 | 57,50 | 64,17 |
| 2019 A | 35,83 | 46,67 | 56,67 |
| 2020 S | 30,83 | 64,17 | 63,33 |
| 2020 A | 30,00 | 53,33 | 56,67 |
| **Endocrinology & diabetology (pediatric)** *Endokrynologia i diabetologia dziecięca* | | | |
| 2018 S | 37,50 | 70,83 | 72,50 |
| 2019 S | 42,50 | 66,67 | 75,00 |
| 2019 A | 42,50 | 67,50 | 66,67 |
| 2020 S | 35,00 | 70,00 | 73,33 |
| 2020 A | 40,83 | 65,83 | 71,67 |
| **Family medicine** *Medycyna rodzinna* | | | |
| 2018 S | 53,33 | 82,50 | 80,83 |
| 2018 A | 53,33 | 74,17 | 77,50 |
| 2019 S | 45,00 | 75,83 | 79,17 |
| 2019 A | 54,17 | 77,50 | 78,33 |
| 2020 S | 52,50 | 75,00 | 81,67 |
| 2020 A | 45,83 | 67,50 | 73,33 |
| **Gastroenterology** *Gastroenterologia* | | | |
| 2018 S | 36,67 | 63,33 | 70,00 |
| 2018 A | 33,33 | 57,50 | 63,33 |
| 2019 S | 35,83 | 64,17 | 66,67 |
| 2019 A | 35,83 | 63,33 | 62,50 |
| 2020 S | 36,67 | 66,67 | 67,50 |
| **Gastroenterology (pediatric)** *Gastroenterologia dziecięca* | | | |
| 2018 S | 32,50 | 61,67 | 62,50 |

**Specialties G-M**

| Exam | Result [%] | | |
|---|---|---|---|
| | gpt-3.5-turbo | gpt-4-0613 | gpt-4-0125-preview |
| **Geriatrics** *Geriatria* | | | |
| 2018 S | 34,17 | 67,50 | 70,00 |
| 2018 A | 44,17 | 73,33 | 81,67 |
| 2019 S | 41,67 | 68,33 | 72,50 |
| 2019 A | 47,50 | 67,50 | 70,83 |
| 2020 S | 41,67 | 65,83 | 74,17 |
| 2020 A | 36,67 | 61,67 | 66,67 |
| **Gynecological endocrinology & reproductive medicine** *Endokrynologia ginekologiczna i rozrodczość* | | | |
| 2018 A | 34,17 | 55,00 | 63,33 |
| 2020 A | 27,50 | 49,17 | 50,00 |
| **Gynecological oncology** *Ginekologia onkologiczna* | | | |
| 2018 S | 40,00 | 56,67 | 62,50 |
| 2018 A | 40,83 | 52,50 | 61,67 |
| 2019 S | 30,83 | 56,67 | 58,33 |
| 2019 A | 33,33 | 65,00 | 62,50 |
| 2020 S | 33,33 | 57,50 | 61,67 |
| 2020 A | 47,50 | 68,33 | 68,33 |
| **Hematology** *Hematologia* | | | |
| 2018 S | 42,50 | 73,33 | 77,50 |
| 2018 A | 33,33 | 65,83 | 68,33 |
| 2019 S | 32,50 | 68,33 | 74,17 |
| 2019 A | 38,33 | 64,17 | 71,67 |
| 2020 S | 45,83 | 70,83 | 77,50 |
| 2020 A | 36,67 | 61,67 | 60,00 |
| **Hypertensiology** *Hipertensjologia* | | | |
| 2018 A | 27,50 | 50,83 | 55,00 |
| 2019 S | 24,17 | 44,17 | 52,50 |
| 2019 A | 49,17 | 70,83 | 66,67 |
| **Infectious diseases** *Choroby zakaźne* | | | |
| 2018 S | 34,17 | 63,33 | 69,17 |
| 2018 A | 31,67 | 69,17 | 74,17 |
| 2019 S | 33,33 | 64,17 | 70,83 |
| 2019 A | 35,83 | 64,17 | 64,17 |
| 2020 S | 37,50 | 62,50 | 73,33 |
| 2020 A | 37,50 | 69,17 | 70,83 |
| **Internal medicine** *Choroby wewnętrzne* | | | |
| 2018 S | 40,83 | 74,17 | 79,17 |
| 2018 A | 37,50 | 73,33 | 75,00 |
| 2019 S | 52,50 | 75,00 | 84,17 |
| 2019 A | 36,67 | 69,17 | 75,83 |
| 2020 S | 45,83 | 75,83 | 78,33 |
| 2020 A | 40,83 | 72,50 | 81,67 |
| **Maxillofacial surgery** *Chirurgia szczękowo-twarzowa* | | | |
| 2018 S | 34,17 | 45,83 | 50,83 |
| 2018 A | 30,83 | 37,50 | 43,33 |
| 2019 A | 34,17 | 46,67 | 52,50 |
| 2020 S | 22,50 | 40,00 | 38,33 |
| 2020 A | 27,50 | 47,50 | 51,67 |
| **Medical rehabilitation** *Rehabilitacja medyczna* | | | |
| 2018 S | 45,00 | 77,50 | 80,00 |
| 2018 A | 40,00 | 67,50 | 73,33 |
| 2019 S | 40,00 | 69,17 | 70,00 |
| 2019 A | 43,33 | 73,33 | 71,67 |
| 2020 S | 45,00 | 67,50 | 70,00 |
| 2020 A | 40,83 | 70,83 | 71,67 |

**Specialties N-O**

| Exam | Result [%] | | |
|---|---|---|---|
| | gpt-3.5-turbo | gpt-4-0613 | gpt-4-0125-preview |
| **Neonatology** *Neonatologia* | | | |
| 2018 S | 29,17 | 62,50 | 61,67 |
| 2018 A | 36,67 | 58,33 | 62,50 |
| 2019 S | 30,83 | 58,33 | 62,50 |
| 2019 A | 30,00 | 59,17 | 60,83 |
| 2020 S | 25,83 | 55,83 | 61,67 |
| 2020 A | 44,17 | 59,17 | 58,33 |
| **Nephrology** *Nefrologia* | | | |
| 2018 S | 36,67 | 67,50 | 70,00 |
| 2018 A | 40,83 | 67,50 | 69,17 |
| 2019 S | 45,00 | 77,50 | 75,83 |
| 2019 A | 33,33 | 71,67 | 65,83 |
| 2020 S | 42,50 | 75,00 | 75,00 |
| 2020 A | 40,83 | 67,50 | 71,67 |
| **Neurology** *Neurologia* | | | |
| 2018 S | 41,67 | 66,67 | 74,17 |
| 2018 A | 48,33 | 71,67 | 71,67 |
| 2019 S | 45,00 | 75,00 | 74,17 |
| 2019 A | 44,17 | 75,83 | 77,50 |
| 2020 S | 40,83 | 71,67 | 71,67 |
| 2020 A | 37,50 | 70,83 | 73,33 |
| **Neurology (pediatric)** *Neurologia dziecięca* | | | |
| 2018 S | 32,50 | 60,00 | 68,33 |
| 2019 S | 43,33 | 69,17 | 75,00 |
| 2019 A | 39,17 | 70,00 | 70,83 |
| 2020 A | 40,00 | 73,33 | 71,67 |
| **Neurosurgery** *Neurochirurgia* | | | |
| 2018 S | 34,17 | 61,67 | 65,00 |
| 2018 A | 38,33 | 55,00 | 63,33 |
| 2019 S | 29,17 | 45,83 | 51,67 |
| 2019 A | 28,33 | 55,83 | 64,17 |
| 2020 S | 30,83 | 60,00 | 56,67 |
| **Obstetrics & gynecology** *Położnictwo i ginekologia* | | | |
| 2018 S | 37,50 | 55,00 | 58,33 |
| 2018 A | 29,17 | 58,33 | 64,17 |
| 2019 S | 25,00 | 54,17 | 56,67 |
| 2019 A | 39,17 | 62,50 | 65,83 |
| 2020 S | 35,00 | 61,67 | 62,50 |
| 2020 A | 38,33 | 57,50 | 63,33 |
| **Occupational medicine** *Medycyna pracy* | | | |
| 2018 S | 40,00 | 58,33 | 56,67 |
| 2018 A | 42,50 | 55,00 | 60,00 |
| 2019 S | 32,50 | 50,83 | 56,67 |
| 2019 A | 32,50 | 51,67 | 46,67 |
| 2020 S | 35,83 | 55,83 | 47,50 |
| 2020 A | 37,50 | 54,17 | 60,00 |
| **Ophthalmology** *Okulistyka* | | | |
| 2018 S | 42,50 | 63,33 | 67,50 |
| 2018 A | 35,00 | 60,83 | 65,00 |
| 2019 S | 42,50 | 65,83 | 69,17 |
| 2019 A | 31,67 | 61,67 | 63,33 |
| 2020 S | 30,83 | 60,00 | 63,33 |
| 2020 A | 45,83 | 70,83 | 67,50 |
| **Orthodontics** *Ortodoncja* | | | |
| 2018 S | 24,17 | 39,17 | 34,17 |
| 2018 A | 31,67 | 41,67 | 38,33 |
| 2019 S | 32,50 | 43,33 | 39,17 |
| 2019 A | 34,17 | 50,00 | 53,33 |
| 2020 S | 31,67 | 52,50 | 47,50 |
| 2020 A | 30,83 | 45,83 | 48,33 |

**Specialties O-P**

| Exam | Result [%] | | |
|---|---|---|---|
| | gpt-3.5-turbo | gpt-4-0613 | gpt-4-0125-preview |
| **Orthopedics** *Ortopedia* | | | |
| 2018 S | 28,33 | 54,17 | 60,83 |
| 2018 A | 30,83 | 60,00 | 63,33 |
| 2019 A | 33,33 | 49,17 | 55,00 |
| 2020 S | 28,33 | 59,17 | 64,17 |
| 2020 A | 24,17 | 55,83 | 64,17 |
| **Otorhinolaryngology** *Otorynolaryngologia* | | | |
| 2018 S | 40,83 | 64,17 | 68,33 |
| 2018 A | 33,33 | 59,17 | 66,67 |
| 2019 A | 28,33 | 53,33 | 58,33 |
| 2020 S | 32,50 | 50,83 | 53,33 |
| 2020 A | 27,50 | 55,00 | 55,00 |
| **Palliative medicine** *Medycyna paliatywna* | | | |
| 2018 S | 45,00 | 72,50 | 76,67 |
| 2018 A | 45,83 | 62,50 | 70,83 |
| 2019 S | 47,50 | 76,67 | 78,33 |
| 2019 A | 35,00 | 63,33 | 64,17 |
| 2020 S | 42,50 | 80,00 | 79,17 |
| 2020 A | 47,50 | 75,00 | 75,00 |
| **Pathology** *Patomorfologia* | | | |
| 2018 S | 42,50 | 68,33 | 76,67 |
| 2018 A | 45,00 | 65,83 | 69,17 |
| 2019 S | 46,67 | 77,50 | 80,83 |
| 2019 A | 38,33 | 73,33 | 79,17 |
| 2020 S | 41,67 | 72,50 | 75,83 |
| 2020 A | 42,50 | 78,33 | 75,00 |
| **Pediatrics** *Pediatria* | | | |
| 2018 S | 35,83 | 70,00 | 68,33 |
| 2018 A | 40,00 | 62,50 | 70,00 |
| 2019 S | 32,50 | 61,67 | 72,50 |
| 2019 A | 39,17 | 68,33 | 70,83 |
| 2020 S | 41,67 | 79,17 | 85,83 |
| 2020 A | 44,17 | 74,17 | 78,33 |
| **Perinatology** *Perinatologia* | | | |
| 2018 A | 35,00 | 63,33 | 64,17 |
| 2020 A | 35,00 | 55,00 | 55,00 |
| **Periodontology** *Periodontologia* | | | |
| 2018 S | 33,33 | 56,67 | 57,50 |
| 2018 A | 34,17 | 59,17 | 63,33 |
| 2019 S | 43,33 | 60,83 | 65,00 |
| 2019 A | 34,17 | 55,83 | 60,00 |
| 2020 S | 36,67 | 59,17 | 56,67 |
| 2020 A | 36,67 | 48,33 | 50,83 |
| **Prosthodontics** *Protetyka stomatologiczna* | | | |
| 2018 S | 35,00 | 56,67 | 54,17 |
| 2018 A | 33,33 | 55,83 | 60,83 |
| 2019 S | 44,17 | 54,17 | 55,00 |
| 2019 A | 41,67 | 50,00 | 55,83 |
| 2020 S | 29,17 | 53,33 | 58,33 |
| 2020 A | 32,50 | 52,50 | 57,50 |
| **Psychiatry (adult)** *Psychiatria* | | | |
| 2018 S | 44,17 | 67,50 | 69,17 |
| 2018 A | 40,83 | 74,17 | 69,17 |
| 2019 S | 44,17 | 75,83 | 75,83 |
| 2019 A | 45,00 | 76,67 | 77,50 |
| 2020 S | 45,00 | 67,50 | 73,33 |
| 2020 A | 45,83 | 72,50 | 77,50 |
| **Psychiatry (child & adolescent)** *Psychiatria dzieci i młodzieży* | | | |
| 2018 S | 45,00 | 67,50 | 71,67 |
| 2018 A | 44,17 | 71,67 | 74,17 |
| 2019 S | 56,67 | 74,17 | 80,00 |
| 2019 A | 45,00 | 73,33 | 73,33 |
| 2020 S | 56,67 | 80,00 | 80,83 |

**Specialties P-V**

| Exam | Result [%] | | |
|---|---|---|---|
| | gpt-3.5-turbo | gpt-4-0613 | gpt-4-0125-preview |
| **Pulmonology** *Choroby płuc* | | | |
| 2018 S | 40,83 | 70,00 | 69,17 |
| 2018 A | 37,50 | 81,67 | 80,83 |
| 2019 S | 36,67 | 70,00 | 73,33 |
| 2019 A | 30,83 | 75,83 | 73,33 |
| 2020 S | 41,67 | 75,00 | 72,50 |
| 2020 A | 35,83 | 69,17 | 66,67 |
| **Pulmonology (pediatric)** *Choroby płuc dzieci* | | | |
| 2020 S | 37,50 | 67,50 | 65,83 |
| 2020 A | 44,17 | 67,50 | 75,00 |
| **Radiation oncology** *Radioterapia onkologiczna* | | | |
| 2018 S | 40,83 | 62,50 | 64,17 |
| 2018 A | 38,33 | 63,33 | 63,33 |
| 2019 S | 35,00 | 63,33 | 63,33 |
| 2019 A | 31,67 | 61,67 | 57,50 |
| 2020 A | 39,17 | 56,67 | 58,33 |
| **Radiology & medical imaging** *Radiologia i diagnostyka obrazowa* | | | |
| 2018 A | 39,17 | 67,50 | 66,67 |
| 2019 S | 40,83 | 67,50 | 74,17 |
| 2019 A | 35,00 | 57,50 | 65,00 |
| 2020 S | 40,00 | 74,17 | 74,17 |
| 2020 A | 14,17 | 18,33 | 15,83 |
| **Rheumatology** *Reumatologia* | | | |
| 2018 S | 31,67 | 66,67 | 65,83 |
| 2018 A | 33,33 | 60,00 | 65,00 |
| 2019 S | 34,17 | 72,50 | 74,17 |
| 2019 A | 35,83 | 66,67 | 65,83 |
| 2020 S | 35,00 | 61,67 | 65,83 |
| 2020 A | 35,00 | 57,50 | 61,67 |
| **Sports medicine** *Medycyna sportowa* | | | |
| 2018 S | 43,33 | 72,50 | 72,50 |
| 2018 A | 46,67 | 74,17 | 76,67 |
| 2019 A | 45,83 | 71,67 | 78,33 |
| **Surgery (general)** *Chirurgia ogólna* | | | |
| 2018 S | 39,17 | 67,50 | 70,83 |
| 2018 A | 40,83 | 69,17 | 69,17 |
| 2019 S | 40,00 | 67,50 | 71,67 |
| 2019 A | 43,33 | 67,50 | 70,83 |
| 2020 S | 41,67 | 71,67 | 75,83 |
| 2020 A | 31,67 | 69,17 | 69,17 |
| **Surgery (pediatric)** *Chirurgia dziecięca* | | | |
| 2018 S | 27,50 | 51,67 | 65,00 |
| 2018 A | 40,00 | 60,00 | 65,83 |
| 2019 S | 33,33 | 62,50 | 68,33 |
| 2019 A | 31,67 | 54,17 | 65,00 |
| 2020 S | 34,17 | 55,00 | 62,50 |
| 2020 A | 40,00 | 64,17 | 63,33 |
| **Surgical oncology** *Chirurgia onkologiczna* | | | |
| 2018 S | 34,17 | 65,00 | 65,00 |
| 2018 A | 30,00 | 70,83 | 70,00 |
| 2019 S | 30,83 | 64,17 | 65,83 |
| 2019 A | 37,50 | 61,67 | 66,67 |
| 2020 S | 32,50 | 59,17 | 58,33 |
| 2020 A | 38,33 | 65,83 | 69,17 |
| **Vascular surgery** *Chirurgia naczyniowa* | | | |
| 2018 S | 34,17 | 55,83 | 67,50 |
| 2018 A | 30,00 | 53,33 | 57,50 |
| 2019 S | 33,33 | 60,00 | 65,00 |
| 2019 A | 34,17 | 53,33 | 56,67 |
| 2020 S | 35,00 | 64,17 | 65,83 |
| 2020 A | 32,50 | 52,50 | 60,00 |

***Table 1*** *GPT models' performance on PES exams. The results for each specialty are presented in alphabetical order. Each specialty's original Polish name is presented in the English translation. "S" stands for the Spring edition of the exam, and "A" stands for the Autumn edition. A gray color indicates a passing score of at least 60%.*

# Results

## Pass rate ratio

**Table 1** displays the exact scores achieved by each model in every exam included in our study. Each PES exam consists of 120 questions with five possible answers each, with only one being correct. That means random guessing results in 20% effectiveness. Doctors must answer 60% of the questions correctly to pass. For this reason, our analysis centered on the GPT models' achievement of surpassing this threshold. The findings were as follows: gpt-3.5-turbo did not pass any exam, gpt-4-0613 passed 184 (62%), and gpt-4-0125-preview passed 222 (75%) of all 297 exams.

## Specialty analysis

In this section, we further explore gpt-4-0613 model's results in relation to specialties due to our confidence in its data integrity, a point we will discuss in more detail in the Discussion section. For this purpose, we excluded specialties with less than four available exams. The gpt-4-0613 failed all the tests for the following specialties: Balneology & physical medicine, Conservative dentistry, Dental surgery, Dentistry (pediatric), Maxillofacial surgery, Occupational medicine, Orthodontics, Prosthodontics. At least one test was passed by the model in the following specialties: Emergency medicine, Endocrinology & diabetology (pediatric), Family medicine, Geriatrics, Hematology, Infectious diseases, Internal Medicine, Medical rehabilitation, Nephrology, Neurology, Neurology (pediatric), Ophthalmology, Orthopedics, Palliative medicine, Pathology, Pediatrics, Psychiatry (adult), Psychiatry (child & adolescent), Pulmonology, Surgery (general).

## Conclusions on GPTs' performance

This study shows that the fourth version of GPT, especially gpt-4-0125-preview, performs well on Polish Board Certification Exams in most specialties. Furthermore, the results indicate a gradual improvement of successive GPT models in correctly answering multiple-choice questions that assess specialist-level medical knowledge and reasoning skills. Another observation is that the results vary significantly depending on the medical domain.

# Discussion

Several considerations should be made regarding GPT's overall performance. As we used OpenAI's API, which does not facilitate real-time internet knowledge access for AI, the experiment provided insights into the internal knowledge and reasoning capabilities of GPT models. This may not be true for the default ChatGPT website. Furthermore, PES utilizes close-end questions with five possible answers. LLMs such as GPT can provide answers based not on their medical knowledge but rather on other features, such as the linguistic structure of answers, bias in question design, etc., since the language models may be better than humans at recognizing these patterns. If the questions were open, the results could differ significantly.

## Possible dataset contamination of gpt-4-0125-preview

Since NIL published the exams on 2023-04-03, they were not used for gpt-3.5-turbo and gpt-4-0613 models training, as their training data is up to September 2021. It does not hold for gpt-4-0125-preview; training data is up to December 2023. It is still unlikely that this last model would encounter the answer keys, especially with relation to original questions, since all files were zipped, every test and its respective key were stored in separate files, and the key file contained only the question number with answer letter and not the full question with contains of each answer. Nevertheless, data contamination with the newest model is still possible. Another aspect is that specific questions and their possible correct answers, even if not word-to-word identical to the official versions, could become objects of online debate among examinees and other doctors even before the official documents were published.

## Reproducibility

In our experiment, the temperature was set to 0. This parameter controls the diversity of the model's output, with the higher values resulting in greater randomness. Even if the temperature is maximally low, there is some minor indeterminism in GPT text generation (https://152334h.github.io/blog/non-determinism-in-gpt-4/), which may possibly cause different answers selected by a model in different runs. This could potentially result in the slightly disparate performance of GPT in specific tests if our experiment was to be replicated using an identical methodology.

## Performance across specialties

The initial analysis of the gpt-4-0613 model's performance reveals areas for further study, especially regarding its underperformance in dental disciplines. The model notably excels in family medicine, a specialty synthesizes knowledge from diverse medical branches yet tends toward a broader generality. This suggests that it's the precision and detail, rather than the scope of required information, that mainly influences gpt-4-0613 model's effectiveness. The ease of access to family medicine information, both in academic and popular formats, might aid this success. The model's performance is less reliable in fields with rapidly changing protocols, possibly because it does not distinguish between newer and older guidelines, treating all available data equally. This behavior points to a gap in GPT's ability to prioritize up-to-date information, a vital skill in accurately responding to specialty-specific questions.

As was mentioned before, it was impossible to automatically retrieve high-quality data from a considerable part of digitalized PES files, which resulted in those exams being excluded from our study at this time. Specialties are not equally represented in the context of the total number of exams included, thus limiting our ability to offer a global comparison of GPT's performance regarding specific PES sections. Even so, it is clear that in some specialties, the tested models achieved much better scores than in others. We presume it is a result of the unequal distribution of content related to each specialty in the datasets used in the training of GPT rather than an indication of how "difficult" a field is.

## Tests and usefulness in real-world applications

Clinical work involves analyzing various types of information and solving complex problems with more than only five possible selected solutions. Doctors need to decide what data they require, skillfully acquire it, and consider the available options in the context of the information gathered, all in order to provide adequate medical help. This includes interviewing patients, performing physical examinations, and ordering diagnostic tests and consultations. It relies significantly on direct human interaction, which no AI model can offer. While GPT's performance in PES is impressive, we want to point out that multiple-choice question tests are only one of many and very limited ways of assessing one's expertise [43]. Thus, the fact that GPT can pass PES does not mean it is a reasonable alternative to a human doctor.

Furthermore, this does not mean that this model could formally become a specialist doctor in Poland. To acquire such a title, a doctor must complete several years of training, attend multiple courses, and perform specific medical procedures. Written PES is not the only step of knowledge verification along the way, as there is also an oral exam.

However, we believe that LLMs' performance in benchmarks such as ours shows its great potential in aiding medical professionals in their duties. While the final medical decision should always be made and authorized by qualified personnel, GAI has many potential utilizations, such as information search and summarization or administrative tasks. The LLMs' clinical application is a current research topic[16]. The role of GAI as a potential assistant for healthcare providers is particularly relevant, considering staff shortages in the Polish healthcare system.

## Future work

We intend to carry out an experiment regarding all 521 tests published by NIL, including the files without a text layer. Such analysis will most likely require the use of an Optical Character Recognition system along with manual human verification. Moreover, our methodology can be expanded to include the ability of GPT to query the internet or other data sources. Currently, the consumer GPT interface allows querying the internet by GPT itself, but the GPT model decides when to query the internet, and this functionality is rarely used. Another possibility is the comparison of GPT's and examinees' performance regarding specific questions and an analysis of the relation between the difficulty of a question and the chance of GPT providing a correct answer.

# Conclusions

In our experiment, we utilized a dataset of 297 tests of the written part of the Polish Board Certification Examination to assess the performance of the three latest GPT models in the role of examinees. We proved that receiving a passing score is within the capabilities of gpt-4-0125-preview regarding 52 of 57 specialties. Even gpt-4-0613, a model that would most definitely not encounter PES questions during its training, passed several tests in 48 specialties. This study underlines the potential for Artificial Intelligence in Poland, particularly advanced LLMs, to revolutionize medical services in Poland, suggesting that in a future, AI may assist accurately healthcare professionals.

**CRediT roles:**

**Jakub Pokrywka** – Conceptualization, Data curation, Formal Analysis, Investigation, Methodology, Project administration, Software, Supervision, Validation, Writing – original draft, Writing – review & editing

**Jeremi I. Kaczmarek** – Investigation, Visualization, Writing – original draft, Writing – review & editing

**Edward J. Gorzelańczyk** – Investigation, Supervision, Writing – original draft

**Acknowledgments:** We would like to thank Piotr Rybak, who pointed out to us the existence of PES exams in a public digitized format. Thanks to this, he inspired us to conduct this research. We also want to thank Anna Konieczka for critically reviewing this article.

**Conflict of interest:** none declared.

**Funding:** Adam Mickiewicz University financed the access to the OpenAI API and partially some of the author's renumeration.

**Declaration of generative AI in scientific writing:** During the preparation of this work the authors used Grammarly and ChatGPT to improve readability and language. After using this tool/service, the authors reviewed and edited the content as needed and take full responsibility for the content of the publication.

**Summary table**

- Assessing GPT models on medical factual knowledge is essential for creating medical AI assistants.

- The GPT models are proven to be effective in English medical tests.
- There are no comprehensive evaluations of GPT models on tests for medical doctors in Polish.
- This study evaluates three GPT models on 297 Polish medical specialty exams.
- The newest GPT4 can pass most of the tests, and their effectiveness relies strongly on the specialty.